\documentclass[10pt,twocolumn,letterpaper]{article}

\usepackage[pagenumbers]{cvpr} 

\definecolor{cvprblue}{rgb}{0.21,0.49,0.74}
\usepackage[pagebackref,breaklinks,colorlinks,allcolors=cvprblue]{hyperref}
\usepackage{multirow}

\def\paperID{00001} 
\def\confName{CVPR}
\def\confYear{2025}

\title{Leadership Assessment in Pediatric Intensive Care Unit Team Training}

\author{Liangyang Ouyang \\ {\tt\small oyly@iis.u-tokyo.ac.jp}\and Yuki Sakai \\{\tt\small sakai-y@iis.u-tokyo.ac.jp}\and Ryosuke Furuta \\{\tt\small furuta@iis.u-tokyo.ac.jp} 
\and Hisataka Nozawa\\
{\tt\small nozawah-ped@h.u-tokyo.ac.jp}
\and Hikoro Matsui \\
{\tt\small hikoromatsui@googlemail.com}
\and Yoichi Sato \\
{\tt\small ysato@iis.u-tokyo.ac.jp}
\and The University of Tokyo
}

\begin{document}
\maketitle
\begin{abstract}

This paper addresses the task of assessing PICU team's leadership skills by developing an automated analysis framework based on egocentric vision. We identify key behavioral cues, including fixation object, eye contact, and conversation patterns, as essential indicators of leadership assessment. In order to capture these multimodal signals, we employ Aria Glasses to record egocentric video, audio, gaze, and head movement data. We collect one-hour videos of four simulated sessions involving doctors with different roles and levels. To automate data processing, we propose a method leveraging REMoDNaV, SAM, YOLO, and ChatGPT for fixation object detection, eye contact detection, and conversation classification. In the experiments, significant correlations are observed between leadership skills and behavioral metrics, i.e., the output of our proposed methods, such as fixation time, transition patterns, and direct orders in speech. These results indicate that our proposed data collection and analysis framework can effectively solve skill assessment for training PICU teams.
\end{abstract}    
\vspace{-5mm}
\section{Introduction}
\label{sec:intro}

Skill assessment is a fundamental task in video understanding. It involves using videos to evaluate skill proficiency by examining human behavior such as actions, decision-making, and interaction patterns. It has numerous applications in sports quality analysis \cite{pirsiavash2014assessing}, manipulation technical evaluation \cite{li2019manipulation,doughty2018s}, and surgical skill assessment \cite{gao2014jhu}.

In clinical training scenarios, skill assessment is used to evaluate doctors' proficiency and teamwork abilities \cite{kim2006pilot,cooper2010rating}.
Traditionally, this assessment is conducted by observers from a third-person perspective, which requires additional human effort and lacks a quantitative understanding of behavioral dynamics.
For example, gaze patterns and eye contact are difficult to perceive from a third-person view, leading to inaccurate evaluations.

\begin{figure}[h]
  \centering
  \includegraphics[width=\linewidth]{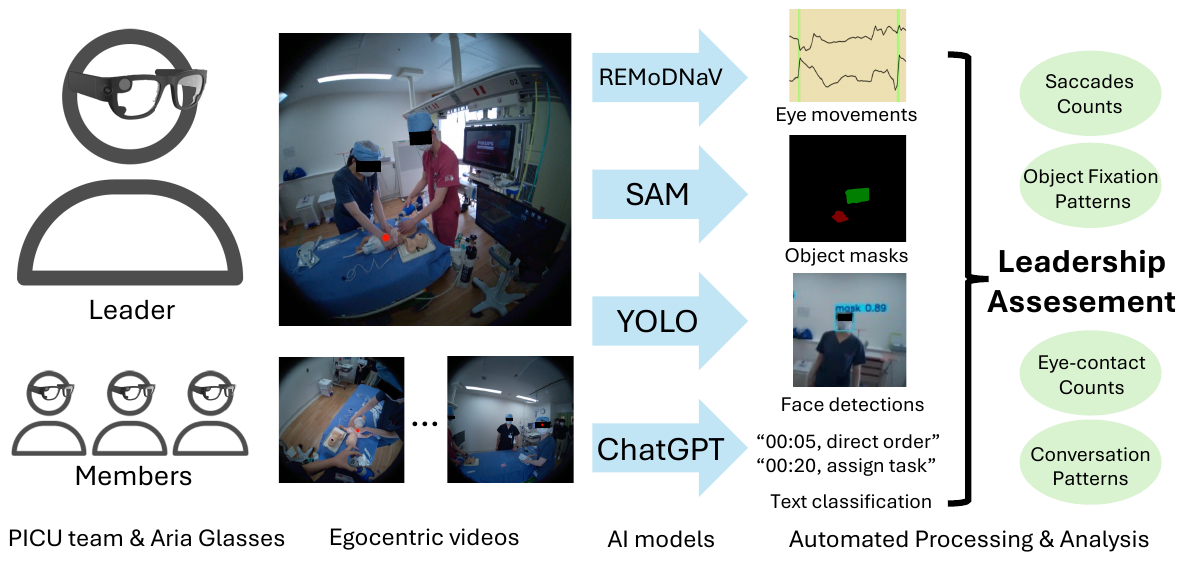}
  \caption{Proposed framework.}
  \vspace{-7mm}
  \label{fig:1_intro}
\end{figure}

To address these challenges, we propose an egocentric vision-based framework for automated skill assessment in clinical training. Recent advancements in wearable devices such as Aria Glasses \cite{engel2023project}, have made it possible to assess skill level using first-person videos \cite{huang2024egoexolearn,wang2023holoassist,grauman2024ego}. These devices also capture multimodal data like gaze and audio, enabling a more comprehensive evaluation of behavioral patterns.

Specifically, we focus on leadership assessment in Pediatric Intensive Care Unit (PICU) team training. As shown in \cref{fig:1_intro}, we collect videos from simulated intensive care sessions in a real hospital PICU. In each session, one leader and three members wear Aria Glasses to capture synchronized egocentric video with other multimodal data. Additionally, we manually annotate conversation transcripts, fixation objects, and leadership scores \cite{cooper2010rating,kim2006pilot} for further analysis.

To assess leadership skills from captured data, we identify key behavioral cues, including fixation object, eye contact, and conversation patterns, as essential indicators of leadership effectiveness. We propose a method leveraging REMoDNaV \cite{dar2021remodnav}, SAM \cite{ravi2024sam2,kirillov2023segment}, YOLO \cite{varghese2024yolov8}, and ChatGPT \cite{openai2025chatgpt} to automate fixation object detection, eye contact detection, and conversation classification. 

Our experiments show a strong correlation between behavioral metrics and leadership skills. For example, experienced leaders exhibit more fixations on patient, eye-contact with team members, and direct orders in speech. This validates our proposed method's efficiency in assessing PICU team leaders' leadership skills.

The main contributions of this work are as follows:
\begin{itemize}
  \item We collect a multimodal dataset of PICU simulated sessions using Aria Glasses. 
  \item We propose an automated analysis framework that integrates REMoDNaV, SAM, YOLO, and ChatGPT to process multimodal data for leadership assessment. 
  \item Our experiments demonstrate that the proposed data collection and analysis framework can efficiently assess the leadership skills of PICU team leaders.
\end{itemize}

\vspace{-4mm}
\section{Related Works}
\vspace{-2mm}
\label{sec:related}

\noindent\textbf{Skill Assessment}
Skill assessment aims to automatically evaluate human performance using behavioral data \cite{grauman2024ego,huang2024egoexolearn,wang2023holoassist}.
Prior research has explored this across domains such as pose estimation in sports \cite{pirsiavash2014assessing}, learned deep features in object manipulation \cite{li2019manipulation,doughty2018s}, and human motion modeling in surgical procedures \cite{gao2014jhu}. 
Our work focuses on assessing leadership skills in PICU team training, where gaze information and communication skills are key indicators.

\noindent\textbf{Egocentric Vision in Clinical Applications}
Egocentric vision has found increasing applications in clinical settings. Previous studies explored wearable assistants \cite{jalaliniya2015designing}, surgery replay systems \cite{matsuda2021surgical}, surgical tool and phase recognition \cite{fujii2024egosurgery}, and skill assessment \cite{gao2014jhu,liu2021towards}. This paper is the first to use multi-person egocentric video for assessing leadership skills in clinical teams.

\vspace{-3mm}
\section{Data Collection}
\vspace{-2mm}
\label{sec:data}
\textbf{Aria Glasses}
Aria Glasses, developed by Meta, is a research device designed for collecting egocentric data to advance AI research in computer vision. It captures multimodal data, including RGB video, audio, eye tracking, SLAM, and IMU signals. In our experiments, we used Profile 15, which records 10 FPS eye-tracking data, 30 FPS RGB frames, and 48 kHz audio.

\noindent\textbf{Session Settings}
We recorded four PICU simulation sessions, each involving four doctors, one facilitator, and one evaluator. All participants are doctors from The University of Tokyo Hospital. The four doctors' roles are leader, nurse, assistant 1, and assistant 2. In each session, a doctor with different levels of clinical experience serves as the leader, who is responsible for assigning orders and tasks. The other three doctors, referred as members, follow the leader's instructions to finish the simulation. 
The facilitator provides essential patient information during the simulation, such as breathing status and heart rate. The evaluator observes the simulation from a third-person perspective and assigns leadership scores based on established standards \cite{kim2006pilot,cooper2010rating}. 
Each session lasts 10 minutes, following a predefined process.

\noindent\textbf{Annotations}
To facilitate leadership assessment, we performed several manual annotations. 
(1) Leadership Scores. The evaluator provides scores for each session leader using the TEAM \cite{cooper2010rating} and Ottawa \cite{kim2006pilot} standards.
(2) Fixation Object Categories. We annotated fixation objects into five categories: patient, member, screen, device, and unknown. This includes 698 fixations.
(3) Object Bounding Boxes. We manually annotated 20 bounding boxes for patients, screens, and members, which are used for prompting the SAM model.
(4) Conversation Transcripts. We used Whisper \cite{radford2023robust} to transcribe conversations in Japanese and manually correct the results, yielding 429 timestamped sentences.
These annotations are currently applied only to sessions 1 and 2, as leaders in these sessions have a notable difference in clinical experience for comparative analysis.
We plan to annotate more sessions for further research. 
\vspace{-3mm}
\section{Proposed Methods}

\begin{figure*}[h]
  \centering
  \includegraphics[width=\linewidth]{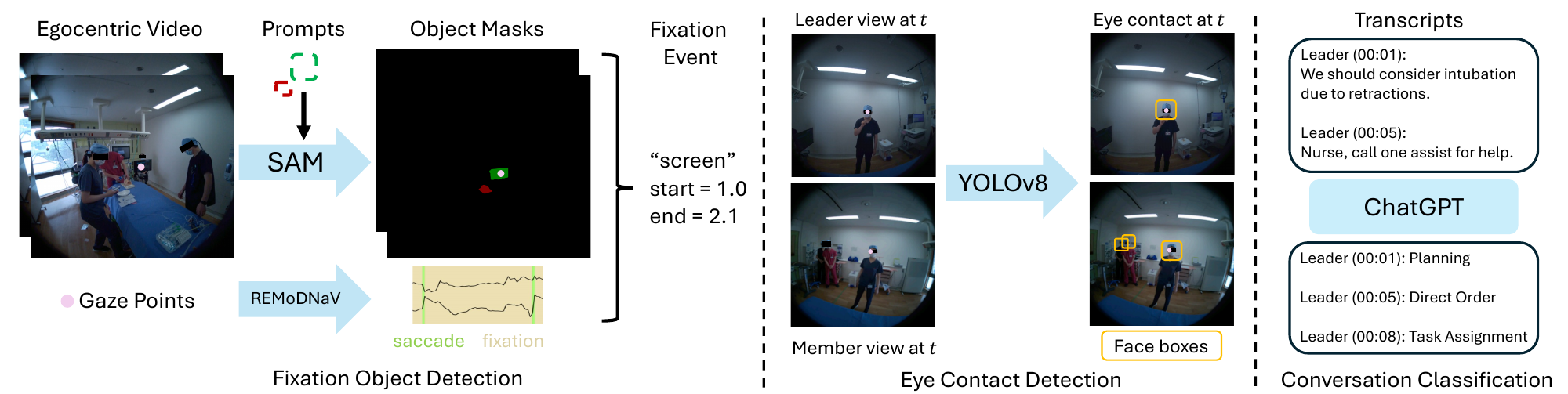}
  \caption{Proposed methods.}
  \vspace{-6mm}
  \label{fig:2_method}
\end{figure*}

\vspace{-1.5mm}
\subsection{Fixation Object Detection}
\vspace{-1.5mm}
Fixation targets can be more accurately identified from egocentric views than third-person perspectives.
In our leadership assessment framework, we include fixation object detection as a key subtask. 
We use REMoDNaV \cite{dar2021remodnav} to analyze eye movement patterns and the Segment Anything Model 2 \cite{ravi2024sam2} to extract segmentation of main objects. 
By combining gaze point with object masks, we detect which object the wearer is focusing on at each fixation event in \cref{fig:2_method}.

Let the egocentric video be denoted as $\mathcal{V} = \{V_t\}^T_{t=1}$, where $V_t \in \mathbb{R}^{H\times W\times 3}$ is the RGB frame at time $t$.
The corresponding gaze point is given by $\mathcal{G} = \{G_t\}, G_t = (h,w).$
We first use REMoDNaV to detect eye movement events

\vspace{-5mm}
\begin{equation}
    \label{eq:rem}
    \centering
    \mathcal{E} =\{(start_i,end_i,E_i)\}_{i=1}^K = REMoDNaV(\mathcal{G})
\end{equation}
\vspace{-5mm}

\noindent where $E_i \in \{0,1,2\}$ denotes the eye movement category, 1 for fixation, 2 for saccade, and 0 for others.
We then use SAM to predict object masks 

\vspace{-4mm}
\begin{equation}
    \label{eq:sam}
    \centering
    \mathcal{M} = \{M_t\}^T_{t=1} = SAM(\mathcal{V},\mathcal{P})
\end{equation}

\noindent where $M_t \in \mathbb{R}^{H\times W}$ is object mask at time $t$. 
Each element $M_t(h,w) \in \{0,1,...,N\}$ denotes the object category at that location.
In our experiments, we choose patient, screen, and three members as $N=3$ main objects. 
Bounding boxes prompts $\mathcal{P}$ are annotated in \cref{sec:data} (3). 

By combining the eye movement and object masks, we identify the fixation object as

\vspace{-6mm}
\begin{equation}
    \label{eq:obj}
    \centering
    O_i = mode(\{M_t(G_t)|start_i\leq t\leq end_i\})
\end{equation}
\vspace{-8mm}

\subsection{Eye Contact Detection}
\vspace{-1.5mm}
Eye contact is a critical non-verbal cue in team collaboration. Experienced leaders tend to engage in timely eye contact with team members to issue instructions and confirmations. In our leadership assessment framework, we detect eye contact events by combining YOLO-detected face bounding boxes with recorded gaze data.

Given the egocentric videos and gaze points of both the leader $(\mathcal{V}_{l},\mathcal{G}_{l})$ and a team member $(\mathcal{V}_{m},\mathcal{G}_{m})$, we use a YOLO \cite{varghese2024yolov8} detector to extract all face bounding boxes in the scene, denoted as $\mathcal{B} = YOLO(\mathcal{V})$.
At timestamp $t$, if both leader and the member's gaze points fall within bounding boxes, we define this as an instance of eye contact.

\vspace{-5mm}
\begin{equation}
    \label{eq:ec}
    \centering
    \small
    EC(t) = \begin{cases}
        1, & G_l(t)~in~B_{m}(t)~\&~G_m(t)~in~B_{l}(t) \\
        0, & otherwise
    \end{cases}
\end{equation}


\subsection{Conversation Classification}
\vspace{-1.5mm}
Following prior work on teamwork evaluation \cite{kolbe2017measuring}, we categorize team leader's conversation into four intent-based classes $C$: \textit{direct order, undirected order, planning, task assignment}. We utilize ChatGPT to classify each leader utterance, allowing for quantitative analysis of leadership communication patterns. This produces a sequence of leader sentences in the form $\mathcal{S}=\{(start_i,end_i,C_i)\}$.

\vspace{-1.5mm}
\subsection{Leadership Assessment}
Based on the evaluation methods in previous research \cite{kim2006pilot,cooper2010rating} and our annotator’s clinical experience, we use the following metrics to assess leadership skills:

\begin{itemize}
  \item \textbf{Object fixation time based on \cref{eq:rem,eq:obj}.} We assume that experienced leaders tend to fixate more on critical objects such as patient and members, rather than on medical devices.
  \item \textbf{Fixation transition patterns based on \cref{eq:rem,eq:obj}.} We suggest that experienced leaders tend to shift their attention across objects in a balanced pattern, as shown by the transition matrix of fixation object sequences.
   \item \textbf{Eye contact counts in \cref{eq:ec}.} We anticipate that experienced leaders engage in more frequent eye contact with team members for non-verbal communication.
  \item \textbf{Ratio of leadership-related sentences in $\mathcal{S}$.} We expect that experienced leaders tend to produce more directive and planning-related utterances.
\end{itemize}

\vspace{-2.5mm}
\section{Experiments}

\subsection{Sub-tasks Results}
\noindent\textbf{Fixation Object Detection.} We evaluated our proposed method for fixation object detection of three categories (patient, members, and screen) on the annotated 698 fixations.
As shown in \cref{tab:fixation}, our method achieves an average accuracy of 88\%, demonstrating its effectiveness in automatically identifying fixations on patient, members and screen.

\noindent\textbf{Eye Contact Detection.} Our method detected 11 eye contact events for Leader 1 and 3 for Leader 2. These eye contacts primarily occurs during orders and confirmations between leader and members, showing the method’s effectiveness in capturing leader behavior relevant to teamwork.

\noindent\textbf{Conversation Classification.}
ChatGPT achieves overall 86.5\% accuracy on 208 utterances of leaders, demonstrating its effectiveness for leadership communication analysis.

\vspace{-3mm}
\begin{table}[h]
\caption{Proposed Methods Accuracy.}
\vspace{-2mm}
\label{tab:fixation}
\centering
\scriptsize
\setlength{\tabcolsep}{1.5mm}{
\begin{tabular}{lccc|ccccc}
\hline
\multirow{2}{*}{Session} & \multicolumn{3}{c}{\underline{Object Fixation Detection}} & \multicolumn{5}{c}{\underline{Conversation Classification}}\\
 & patient & member & screen & DO & UO & PL & TA & all \\ \hline
Leader 1 &  81.9  & 85.6 &  96.1 & 99.1 & 88.5 & 92.9 & 100 & 81.4\\ 
Leader 2 &  90.2  & 87.4 &  89.1 & 97.9 & 94.7 & 97.9 & 100 & 92.6\\ 
\hline
\end{tabular}}
\end{table}
\vspace{-5mm}

\subsection{Leadership Assessment Results}
\noindent\textbf{Fixation Time \& Eye Contact Count \& Conversation Ratio}
As shown in \cref{tab:leader-compare}, we compare two leaders based on proposed quantitative metrics from human annotations.
Leader 1 focuses longer on the patient and team members than on devices, which aligns with expert doctor's experiences.
We also observed that only 4\% of Leader 1’s fixations are unknown, compared to 24\% for Leader 2, suggesting that experienced leaders tend to fixate more on key objects.
Besides, Leader 1 exhibits significantly more eye contact with team members, showing that effective leadership involves greater non-verbal engagement with the team. 
In terms of conversation, Leader 1 use more direct orders, while both leaders have similar levels of undirected orders. 
These results suggest that experienced leaders are better at delivering clear instructions and coordinating the team.
\vspace{-3mm}

\begin{table*}[h]
\caption{Comparison of two leaders on proposed leadership assessment metrics.}
\centering
\small
\setlength{\tabcolsep}{1.5mm}{
\begin{tabular}{@{}lcc|ccccc|c|c@{\hspace{6mm}}c@{\hspace{7mm}}cc@{}}
\toprule
\multirow{2}{*}{Session} & \multicolumn{2}{c}{\underline{Human Evaluation}} & \multicolumn{5}{c}{\underline{Average Object Fixation Time (s)}} & \multicolumn{1}{c}{\underline{EC Count}} & \multicolumn{4}{c}{\underline{Conversation Category Ratio (\%)}} \\
& TEAM \cite{cooper2010rating} & Ottawa \cite{kim2006pilot} & Patient & Member & Screen & Device & Unknown &  & DO & UO & PL & TA \\
\midrule
 Leader 1 & \bf{27/44} & \bf{35/35} & \bf{2.28} & \bf{1.98} & 1.54 & 1.80 & 1.04 & \bf{11} & \bf{17.2} & 21.6 & \bf{11.3} & \bf{3.4}\\
Leader 2 & 12/44 & 26/35 & 1.50 & 1.24 & \bf{2.45} & \bf{2.08} & \bf{1.21} & 3 & 5.3 & \bf{22.1} & 9.5 & 1.1\\
\bottomrule
\end{tabular}}
\label{tab:leader-compare}
\vspace{-6mm}
\end{table*}

\begin{figure}[h]
  \centering
  \includegraphics[width=\linewidth]{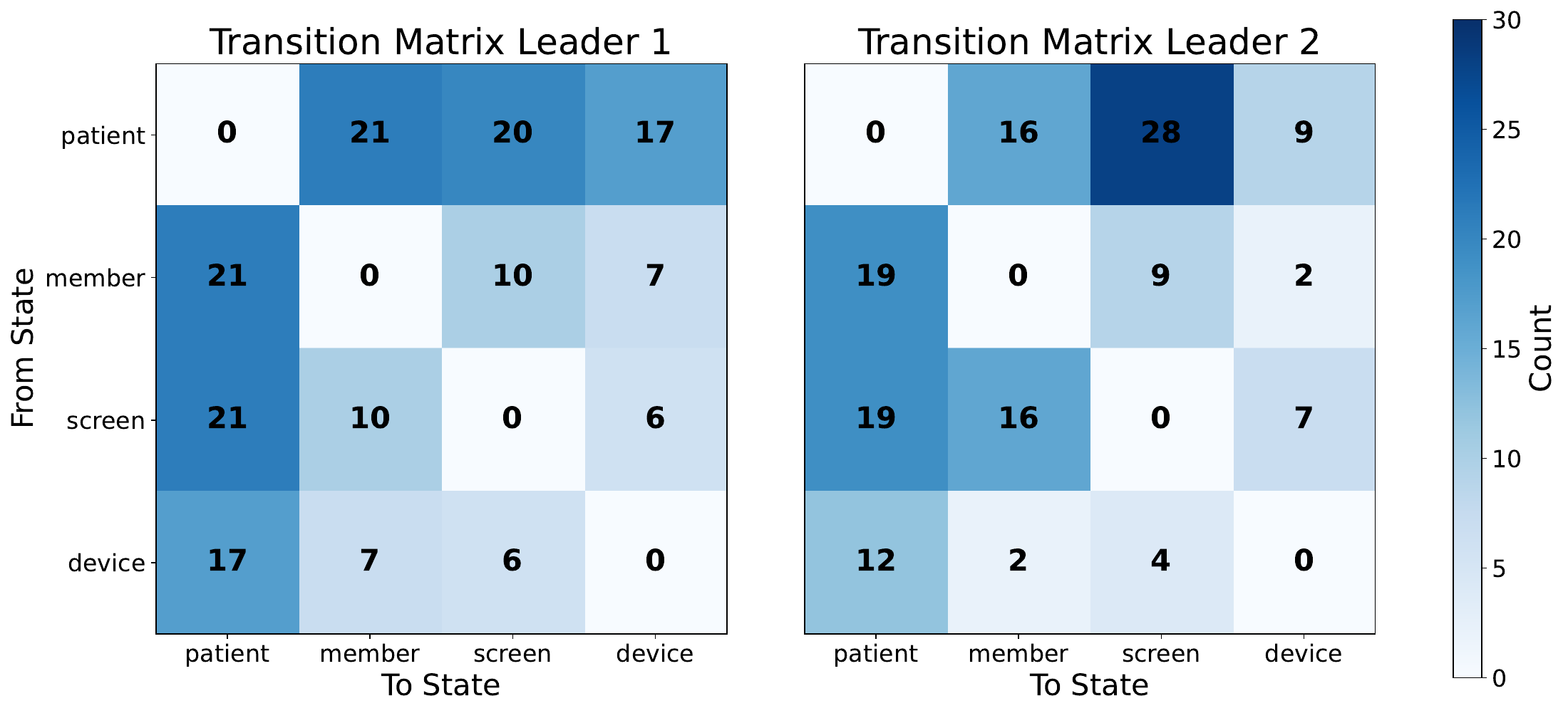}
  \caption{Fixation Transition Matrix.}
  \label{fig:4_matrix}
\end{figure}
\vspace{-4mm}

\noindent\textbf{Fixation Transitions} We represent fixation objects as a sequence of states and construct transition matrices in \cref{fig:4_matrix}.
Leader 1’s transition matrix exhibits strong symmetry, where count of transition A-B is equal to B-A. This balanced fixation transition pattern reflects more structured attention strategies and stronger leadership skills.

\begin{figure}[h]
  \centering
  \includegraphics[width=\linewidth]{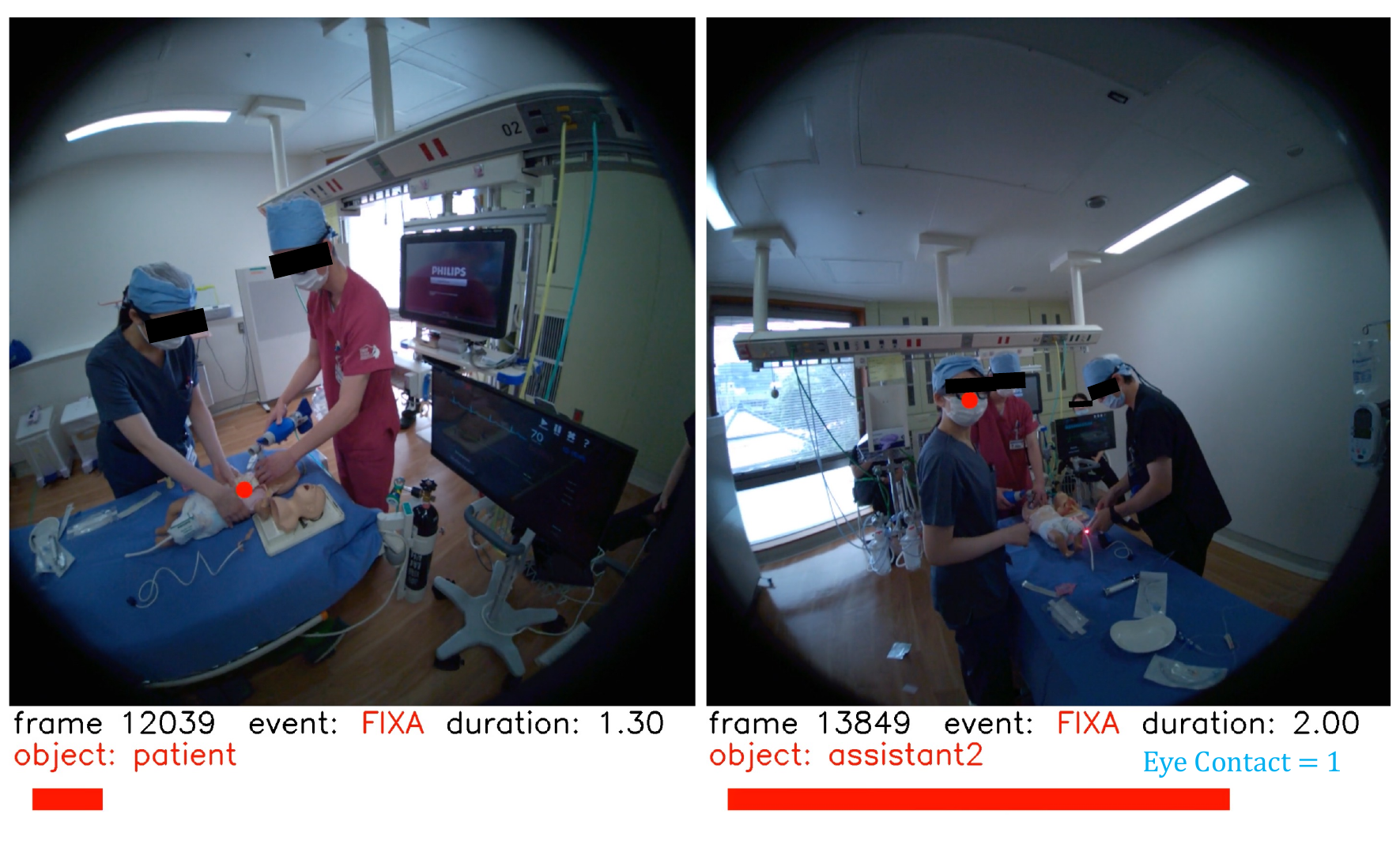}
  \vspace{-10mm}
  \caption{Frames in Visualization Video.}
  \label{fig:3_vis}
  \vspace{-6mm}
\end{figure}

\noindent\textbf{Video Visualization}
We visualized all behavioral metrics on egocentric videos, generating intuitive visualization video clips. 
As shown in \cref{fig:3_vis}, these videos allow doctors to clearly observe leader behaviors, supporting evaluation and educational purposes in clinical applications. 
\vspace{-4mm}
\section{Conclusion}
\vspace{-2mm}
This paper proposes an egocentric vision based framework for assessing leadership skills in PICU team training. 
Using Aria Glasses and a set of AI models, we analyzed key behavioral metrics. 
Our results show that the proposed framework effectively captures differences in fixation objects, eye contact, and communication patterns. 
Current limitations include the data size and reliance on manual annotations. 
We aim to extend this work to broader clinical scenarios, scale up data collection and annotation, and further automate the analysis pipeline.
\vspace{-2mm}
\section*{Acknowledgements}
\vspace{-2mm}
{\spaceskip=0.16em This work is supported by JST ASPIRE Grant Number JPMJAP2303. We appreciate the collaboration of The University of Tokyo Hospital PICU team.} 
\vspace{-3mm}
{
    \small
    \bibliographystyle{ieeenat_fullname}
    \bibliography{main}
}


\end{document}